# Multi-Label Classification Framework for Hurricane Damage Assessment

Zhangding Liu[1], Neda Mohammadi[2], John E. Taylor[3*]

[1]PhD Student, School of Civil and Environmental Engineering, Georgia Institute of Technology, 790 Atlantic Dr NW, Atlanta, GA 30332, United States; e-mail: zliu952@gatech.edu
[2]School of Civil and Environmental Engineering, Georgia Institute of Technology, 790 Atlantic Dr NW, Atlanta, GA 30332, United States; e-mail: nedam@gatech.edu; The University of Sydney, Camperdown NSW 2050, Australia; e-mail: neda.mohammadi@sydney.edu.au
[3]Professor, School of Civil and Environmental Engineering, Georgia Institute of Technology, 790 Atlantic Dr NW, Atlanta, GA 30332, United States; e-mail: jet@gatech.edu (corresponding author)

**ABSTRACT**

Hurricanes cause widespread destruction, resulting in diverse damage types and severities that require timely and accurate assessment for effective disaster response. While traditional single-label classification methods fall short of capturing the complexity of post-hurricane damage, this study introduces a novel multi-label classification framework for assessing damage using aerial imagery. The proposed approach integrates a feature extraction module based on ResNet and a class-specific attention mechanism to identify multiple damage types within a single image. Using the Rescuenet dataset from Hurricane Michael, the proposed method achieves a mean average precision of 90.23%, outperforming existing baseline methods. This framework enhances post-hurricane damage assessment, enabling more targeted and efficient disaster response and contributing to future strategies for disaster mitigation and resilience.

**INTRODUCTION**

Hurricanes are among the most destructive natural disasters, causing widespread infrastructure damage, displacing populations, and leading to significant economic losses (Balaguru et al. 2023). Between 1900 and 2017, the United States experienced 197 hurricanes with 206 landfalls, resulting in nearly $2 trillion in damages when adjusted to 2018 economic conditions (Weinkle et al. 2018). As climate change progresses and coastal populations grow, hurricane-related losses are expected to increase, underscoring the need for automated tools to assess post-disaster infrastructure damage (Jing et al. 2024).

Timely and accurate classification of post-hurricane damage is vital for coordinating emergency responses, prioritizing resource allocation, and facilitating rapid recovery (Al Shafian and Hu 2024; Kaur et al. 2021). Quick identification of severely impacted areas guides rescue operations to ensure aid reaches the most vulnerable populations (Chou et al. 2017). A detailed understanding of damage extent and type is essential for infrastructure repair and reconstruction planning (Liu et al. 2022). However, traditional damage assessment methods rely on ground surveys and manual inspections, which are time-consuming, labor-intensive, and pose risks to rescue teams (Cheng et al. 2021). Similarly, sensor-based numerical simulations depend on pre-installed networks, which are often unavailable in disaster zones and lack the ability to provide large-scale assessment (Lozano and Tien 2023).

Advancements in remote sensing and computer vision have enhanced the potential for automated damage classification. Techniques like satellite imagery and UAV aerial photography have been employed to assess damage, with machine learning algorithms demonstrating capabilities in differentiating various damage types (Wang et al. 2022). Cao and Choe (2020) used CNNs to annotate building damage from satellite imagery after Hurricane Harvey, while McCarthy et al. (2020) showed that machine learning could identify healthy versus degraded vegetation using multispectral images from Hurricane Irma. Khajwal et al. (2023) improved damage prediction accuracy with a multi-view CNN, achieving 65% for precise states and 81% with slight deviations allowed. However, most existing methods rely on single-label classification, which is insufficient to represent the complex contents of real-world images where multiple types of damage may coexist.

Multi-label classification offers superior descriptive capabilities by identifying multiple damage types within a single image, providing a more comprehensive assessment. This approach is more challenging due to the need to recognize various damages that can vary in location, scale, and may overlap or be partially obscured (Zhou et al. 2023). To address these challenges, we propose a novel framework designed for multi-label post-hurricane image classification. Our main contributions are as follows: (i) We propose a novel framework that combines ResNet for feature extraction and a residual attention mechanism to adaptively calibrate channel responses, enabling efficient classification of damage across varying scales and categories; and (ii) Comprehensive experiments on the Rescuenet dataset demonstrate that our method achieves superior accuracy in classifying complex multi-label images, outperforming existing models.

**METHODOLOGY**

This section introduces a novel framework for multi-label post-hurricane image classification, which mainly consists of a feature extraction module and a class-specific residual attention (CSRA) module. The overview of the proposed multi-label damage classification framework is shown in Figure 1.

**Flowchart of the proposed approach.** The pipeline begins with preprocessing and augmenting post-hurricane images, followed by inputting them into the model. We utilize a feature extraction module to get feature matrix $X \in \mathbb{R}^{d \times h \times w}$, where $d$, $h$, and $w$ correspond to the dimensionality, height, and width of the feature map, respectively. CSRA enhances class-specific feature representation by calculating spatial attention scores for each class, which are then integrated with the average pooling features. Residual attention is applied to every score tensor to produce different logits, which are then fused to get the final multi-label prediction.

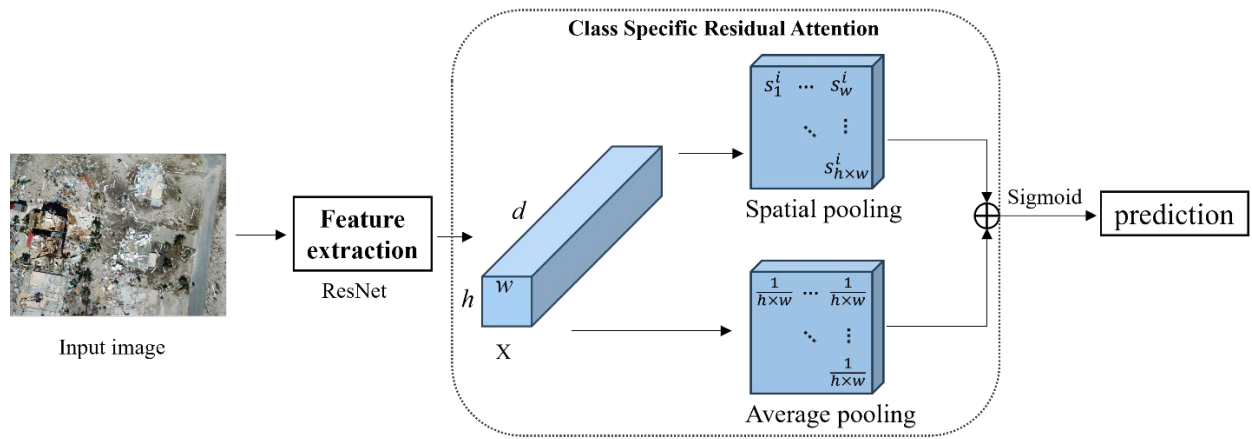

**Figure 1. Overview of the proposed multi-label damage classification framework.**

**Feature extraction module.** With the advancement of deep learning, numerous image classification models have emerged in the field of computer vision. ResNet (He et al. 2015) stands out for its versatile architecture, supporting models with depths ranging from ResNet-18 to ResNet-152. To tackle the vanishing gradient problem in deep networks, ResNet incorporates residual units featuring skip connections. These connections bypass intermediate layers by directly linking the input to the output, ensuring smooth information propagation throughout the network. This innovative approach not only combats the performance degradation associated with increased depth but also enables the construction of significantly deeper models, such as ResNet-50, ResNet-101, and ResNet-152, as illustrated in Figure 2(a).

ResNet-50 consists of four main convolutional stages, each stage is built with bottleneck blocks arranged in a sequence described by k = [3,4,6,3]. In comparison, ResNet-101 follows the same structural design but includes a higher number of blocks, with k = [3,4,23,3]. Both models commence with an initial 7×7 convolutional layer followed by max-pooling, after which the data progresses through the network's stages. Each residual block within these stages includes three convolutional layers—1×1, 3×3, and 1×1—interleaved with batch normalization and ReLU activations. The skip connections in these blocks help the model learn identity mappings, which are crucial for maintaining performance as the network's depth increases. Figure 2(b) illustrates the detailed layout of a typical residual block within these stages.

**Class specific residual attention module.** Some researchers have applied attention mechanisms to improve aerial image classification performance (Hua et al. 2020; Wang et al. 2021). However, few studies have explored the use of attention mechanisms in multi-label post disaster classification scenarios. To enhance classification in multi-label post-disaster scenarios, we utilize a CSRA module to integrate spatial attention. The CSRA module leverages spatial attention to create category-aware features for each class. Specifically, it computes a spatial attention score for each class, which highlights the most relevant regions in the image associated with that class. This attention score is then combined with the class-agnostic global average pooling feature to produce a refined representation (Zhu and Wu, 2021). By integrating both localized and holistic features, CSRA effectively captures the complex spatial dependencies present in post-disaster imagery.

To generate the final predictions, the CSRA module outputs are passed through a fully connected layer equipped with a sigmoid activation function. Unlike the softmax function, which is commonly used for single-label classification, the sigmoid function enables independent prediction for each class by mapping the network's outputs to probabilities between 0 and 1. A threshold of 0.5 is applied to these probabilities to derive the final multi-label predictions: outputs greater than or equal to 0.5 are classified as 1, indicating the presence of the corresponding class, while outputs below 0.5 are classified as 0, indicating its absence. This results in a binary vector like [1, 0, 1, …, 0] to represent class presence in the image.

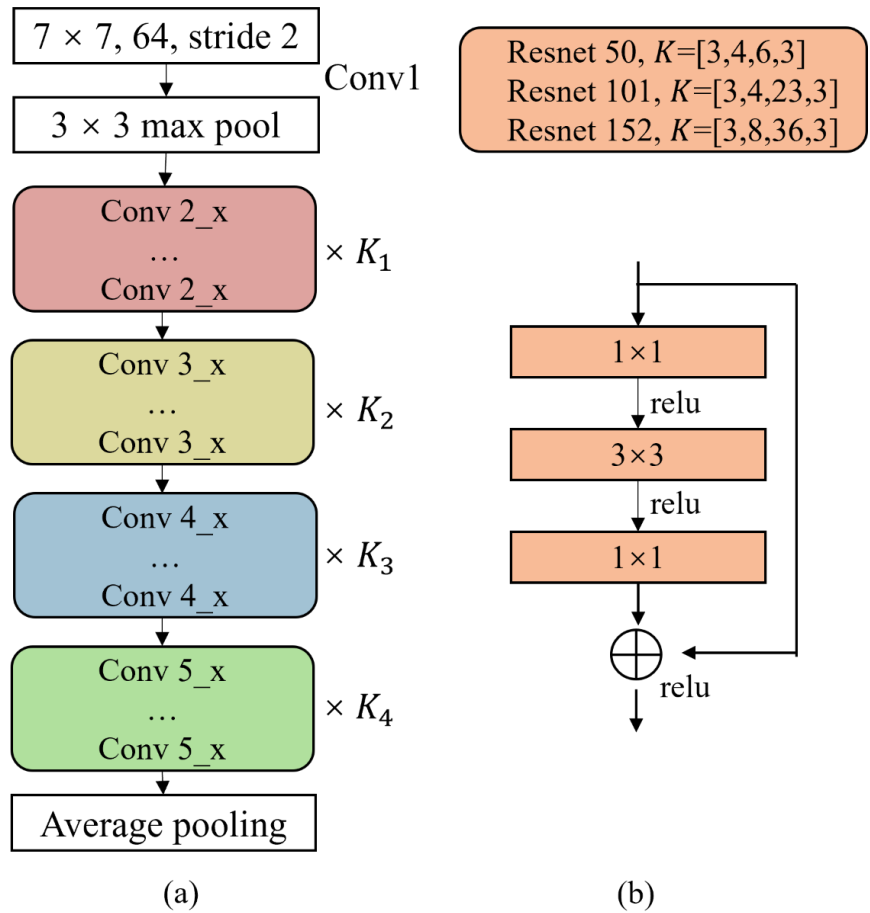

Figure 2: (a) Resnet architecture. (b) Residual block structure.

**Data.** Hurricane Michael struck near Mexico Beach, Florida, on October 10, 2018, causing over $25 billion in damages and 16 fatalities (Beven et al. 2019). The extensive damage caused by Hurricane Michael underscores the importance of accurate post-disaster damage assessment for effective recovery and mitigation planning. This study utilized aerial imagery and labels from the RescueNet dataset (Rahnemoonfar et al. 2023), which consists of a total of 4,494 images. The dataset includes six instance types: road, tree, building, water, vehicle, and pools. Building damage is categorized into four levels: No Damage, Medium Damage, Major Damage, and Total Destruction, while road damage is classified as Clear or Blocked. Several example images and corresponding multi labels from the RescueNet dataset are shown in Figure 3.

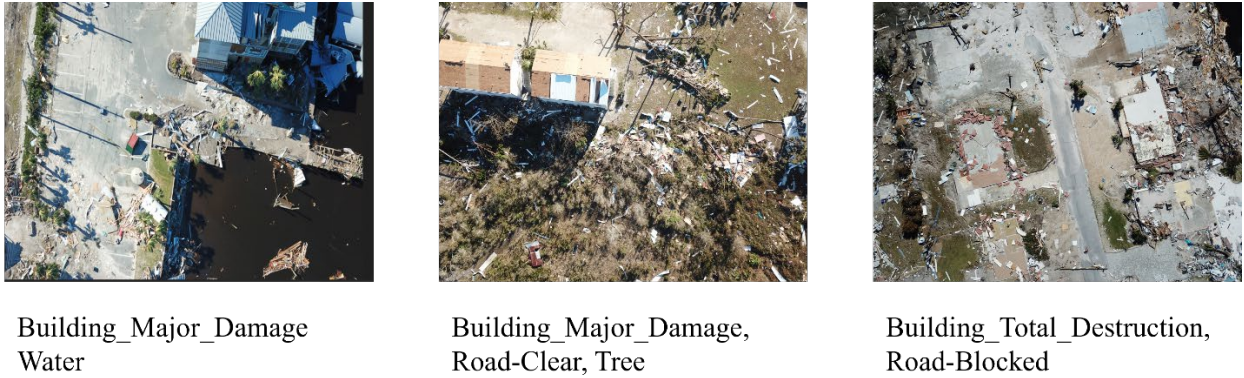

Building_Major_Damage, Water

Building_Major_Damage, Road-Clear, Tree

Building_Total_Destruction, Road-Blocked

**Figure 3. Example images and multi labels of Rescuenet dataset.**

## EXPERIMENT AND RESULTS

**Implementation Details.** To establish effective model performance, a series of comparative experiments are implemented: (i) compare the performance of different feature extraction modules and (ii) investigate the impact of incorporating class aware attention mechanisms on multi-label hurricane damage image classification. In each of our experiments, we divide the RescueNet dataset into three subsets with a ratio of 80% for training, 20% for testing. For the CSRA module and classifiers, a learning rate of 0.1 is selected, while a lower rate of 0.01 is set for the feature extraction module. The feature extraction module is initialized from different pretrained models and trained for 50 epochs on Rescuenet datasets.

**Evaluation Metrics.** The model's classification performance was evaluated using four metrics: mean average precision (mAP), overall precision (OP), overall recall (OR), and overall F1-score (OF1). For each category, Average Precision (AP) evaluates how well the predicted labels align with the actual labels, approximating the area under the precision-recall curve. Mean Average Precision (mAP) is the average AP across all categories and is commonly used to assess overall model performance. Overall Precision (OP) measures the ratio of correctly predicted labels to the total predicted labels, while Overall Recall (OR) represents the proportion of correctly predicted labels relative to all ground truth labels. Overall F1-score (OF1) is the harmonic mean of precision and recall, balancing the trade-off between these metrics.

**Results and Analysis.** This section presents and analyzes the results of the experiments. Table 1 provides a comprehensive summary of the findings, including the performance of different feature extraction models and the impact of incorporating the CSRA module. These results offer valuable insights into the design and optimization of model architectures for a multi-label damage classification framework.

Table 1. The performances of different models. The best performance in each column is indicated by the bold font.

| Method | mAP | OP | OR | OF1 |
|---|---|---|---|---|
| Resnet50 | 86.55 | 85.72 | 78.68 | 82.05 |
| Resnet101 | 87.24 | 84.87 | 81.55 | 83.18 |
| Resnet152 | 88.19 | 85.61 | 80.04 | 83.05 |
| VGG19 | 85.08 | 81.92 | 79.14 | 80.50 |
| EfficientNet | 86.97 | 83.21 | 80.12 | 81.63 |
| ViT-B16+CSRA | 87.03 | 82.96 | 79.99 | 81.45 |
| Resnet50+CSRA | 89.64 | 86.51 | 81.55 | 83.96 |
| Resnet101+CSRA | 89.81 | 86.86 | **82.72** | **84.74** |
| **Resnet152+CSRA** | **90.23** | **87.37** | 81.62 | 84.40 |

First, the performance of various feature extraction models, including ResNet, VGG, and EfficientNet, is compared in Table 1. Deeper networks like ResNet152 outperform shallower ones like ResNet50 due to their larger receptive fields, enabling better feature extraction from complex scenes. Among these feature extraction modules, ResNet152 achieves a notable mAP of 88.19%, demonstrating its advantage in balancing depth and feature extraction ability. Furthermore, integrating the CSRA module significantly enhances model performance. For instance, adding CSRA to ResNet101 increases its mAP from 87.24% to 89.81% and its OF1 from 83.18% to 84.74%. Adding CSRA to ResNet50 increases its mAP from 86.55% to 89.64% and its OF1 from 82.05% to 83.96%. CSRA improves the model's ability to focus on important regions, leading to more accurate predictions. Among all models, the ResNet152+CSRA model achieves the highest mAP of 90.23% and an OP of 87.37%. The combination of ResNet152 and CSRA allows for more accurate classification of post-hurricane damage across varying scales and locations.

**DISCUSSION**

The proposed multi-label classification approach significantly improves the efficiency and comprehensiveness of post-hurricane damage assessment. Disaster environments often involve overlapping structures and occluded features caused by debris and infrastructural collapse. By extracting multi-scale features and capturing class-specific spatial information, the method ensures accurate classification even under challenging conditions. It can concurrently identify various damage types, such as buildings, roads, trees, and water bodies, retaining critical information for targeted recovery plans. Our proposed class-specific attention network addresses these challenges by integrating ResNet for feature extraction and a class-specific residual attention module to refine spatial focus. This fine-grained classification supports effective resource allocation and disaster mitigation while accelerating the transition from data collection to actionable insights.

Despite its advantages, the method has limitations. Our study mainly demonstrates that MLCSANet achieves high overall performance, as evidenced by mAP and OPmetrics. Future work will further evaluate per-class accuracy and explore hierarchical classification for better distinction of intermediate damage levels. Additionally, disaster datasets sometimes are imbalanced, often skewing classification results toward undamaged areas. While this study primarily introduces a new classification framework without extensive data augmentation or hyperparameter tuning, future improvements will focus on dataset balancing, advanced augmentation (e.g., synthetic oversampling), adaptive thresholding strategiesand, and adaptive loss functions to enhance accuracy and effiency across all damage categories.

## CONCLUSION

This study introduced a novel framework for multi-label post-hurricane damage classification. By combining ResNet's feature extraction ability with the class-specific residual attention mechanism, the proposed method effectively identifies diverse damage types within a single image. It outperformed baseline models, achieving a mAP of 90.23%. These findings demonstrate this method's potential as a robust and efficient tool for post-disaster assessment, supporting rapid and comprehensive decision-making in emergency response and resource allocation.